\documentclass[10pt,twocolumn,letterpaper]{article}

\usepackage[pagenumbers]{cvpr}      

\usepackage{microtype}
\usepackage{tabularx}
\renewcommand{\paragraph}[1]{\vspace{.5em}\noindent\textbf{#1.}}

\definecolor{cvprblue}{rgb}{0.21,0.49,0.74}
\usepackage[pagebackref,breaklinks,colorlinks,allcolors=cvprblue]{hyperref}

\def\paperID{398} 
\def\confName{3DV\xspace}
\def\confYear{2027\xspace}

\title{SparseTalk - Sparsifying 3D Gaussian Language Fields for Efficient 3D Visual Question Answering}

\author{
Davit Soselia \qquad Joseph JaJa \qquad Amitabh Varshney\\
University of Maryland, College Park, MD, USA\\
{\tt\small dsoselia@umd.edu \quad josephj@umd.edu \quad varshney@umd.edu}
}

\begin{document}
\maketitle
\begin{abstract}

3D Gaussian language fields provide an explicit, spatially grounded representation for 3D visual question answering (VQA), but their dense semantic features can require tens of thousands of embeddings per scene, resulting in substantial storage, memory, and inference costs. We investigate how much of this representation is actually necessary for downstream reasoning. Starting from a full embedding representation, we systematically sparsify its semantic embeddings, including the previously underexplored regime below a single image-equivalent block down to 8 visual tokens. We compare random, geometric, semantic, and joint spatial-semantic selection strategies and introduce an object-based sparsification method that distributes the token budget across detected object instances while retaining background context. Experiments on ScanQA and MV-ScanQA reveal substantial redundancy in dense Gaussian language fields. Strong VQA performance is retained with only a few hundred semantic embeddings, corresponding to less than 1\% of the original representation. Object-based selection performs well relative to others, with only modest
observed changes down to 256 tokens. At this budget, SparseTalk retains 0.80\% of SplatTalk’s 32,076-token inference input and 0.332\% of the mean 77,207-Gaussian dense field, increasing inference throughput while reducing decoded-feature memory 125-fold. 

\end{abstract}
\section{Introduction}
\label{sec:intro}

As vision-language models are increasingly used in many tasks, including
robotics, physical AI, augmented reality, autonomous systems, and industrial environments,
it has become increasingly important for these models to understand, identify
objects in, and reason about the 3D world. A direct approach is to provide a
sequence of images to a vision-language model (VLM), effectively treating
different observations of a scene as frames of a video. While this approach
benefits from strong pretrained image and video models, it requires the model
to infer persistent 3D structure from a collection of perspective-dependent
2D observations. Existing VLMs continue to struggle with cross-view
integration and spatial relationships between objects, particularly when the
answer requires a world-centric rather than camera-centric understanding of the scene~\cite{hong20233dclr,chen2024spatialvlm,zhang2026multiviewbench}.

\begin{figure}[h]
    \centering
\includegraphics[width=1\linewidth]{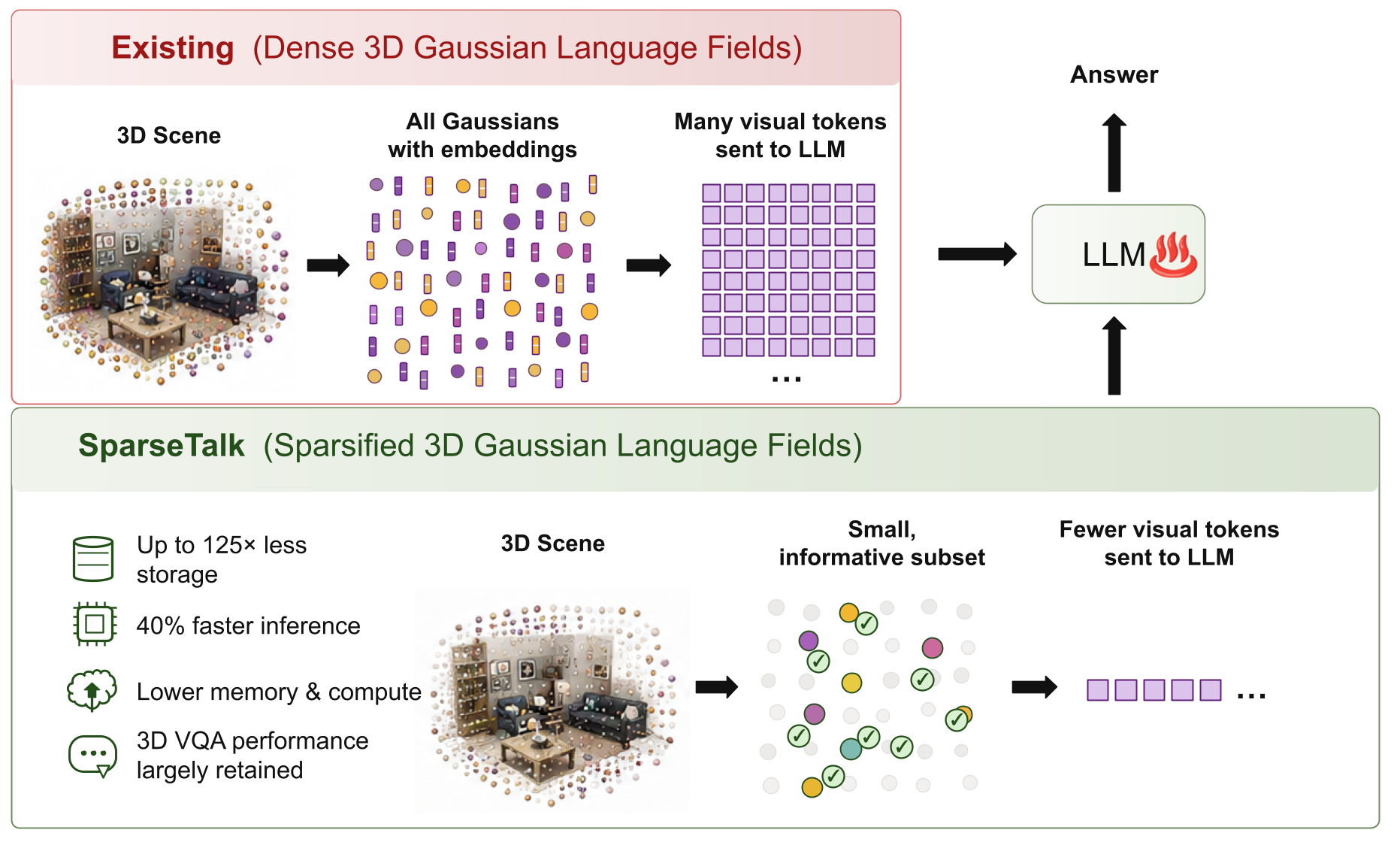}
    \caption{By using object-based sparsification, we reduce the number of semantic embeddings from tens of thousands to a few hundred, leading to faster inference and less memory usage, while retaining most of the VQA performance across the ScanQA and MV-ScanQA datasets.}
    \label{fig:pipeline}
\end{figure}

An alternative direction is to encode semantic information directly within a
3D representation. Methods such as LangSplat, ChatSplat, and
SplatTalk~\cite{qin2024langsplat,chen2024chatsplat,thai2025splattalk} associate
learned language or vision-language features with the primitives of a 3D
Gaussian Splatting representation~\cite{kerbl20233dgs}. These methods often
use an encoder or autoencoder to compress high-dimensional visual features,
then train the Gaussian representation so that rendered feature maps align
with features extracted from the original views. These methods produce a spatially
grounded semantic representation that can support open-vocabulary querying,
conversation, or 3D visual question answering.

However, this representation also introduces a substantial computational
cost. A reconstructed scene may contain tens of thousands of Gaussians,
with a separate semantic embedding associated with each primitive. These
embeddings increase storage, memory use, transfer bandwidth, and the number of
visual tokens that must be processed by the language model. Moreover, even
simple objects may be represented by hundreds of nearby
Gaussians. It is unlikely that every one of these embeddings provides unique
semantic information. Although prior work has compressed the dimensionality
of Gaussian features and compared several token-selection strategies, the
amount of redundancy in these representations, particularly below one
image-worth of tokens, remains underexplored.

Our contributions are threefold. First, we present a controlled study of question-independent, post-hoc subset sparsification for frozen 3D Gaussian language fields, comparing random, geometric, semantic, and joint spatial-semantic selection across budgets from 729 down to 8 retained visual tokens on ScanQA and MV-ScanQA. Second, we introduce an object-based selector that associates multi-view instance masks with 3D Gaussians and constructs a token ranking across foreground-object tracks while preserving a contextual background pool. Third, we characterize the resulting quality-efficiency trade-off using standard metrics and Scaled Visually Attributable Performance (SVAP), a blind-adjusted measure of sparse-model performance relative to full-context performance. Object-based selection achieves the strongest observed aggregate performance among the evaluated selectors, while uniform random sampling provides a surprisingly strong baseline. At \(k=256\), SparseTalk retains only \(0.80\%\) of SplatTalk's 32,076-token inference input, reduces decoded-feature memory from 229.92 MB to 1.84 MB, and increases measured throughput from 0.58 to 14.3 questions per second, with only modest observed changes in answer quality.

\begin{figure*}[th!]
    \centering
\includegraphics[width=0.9\linewidth]{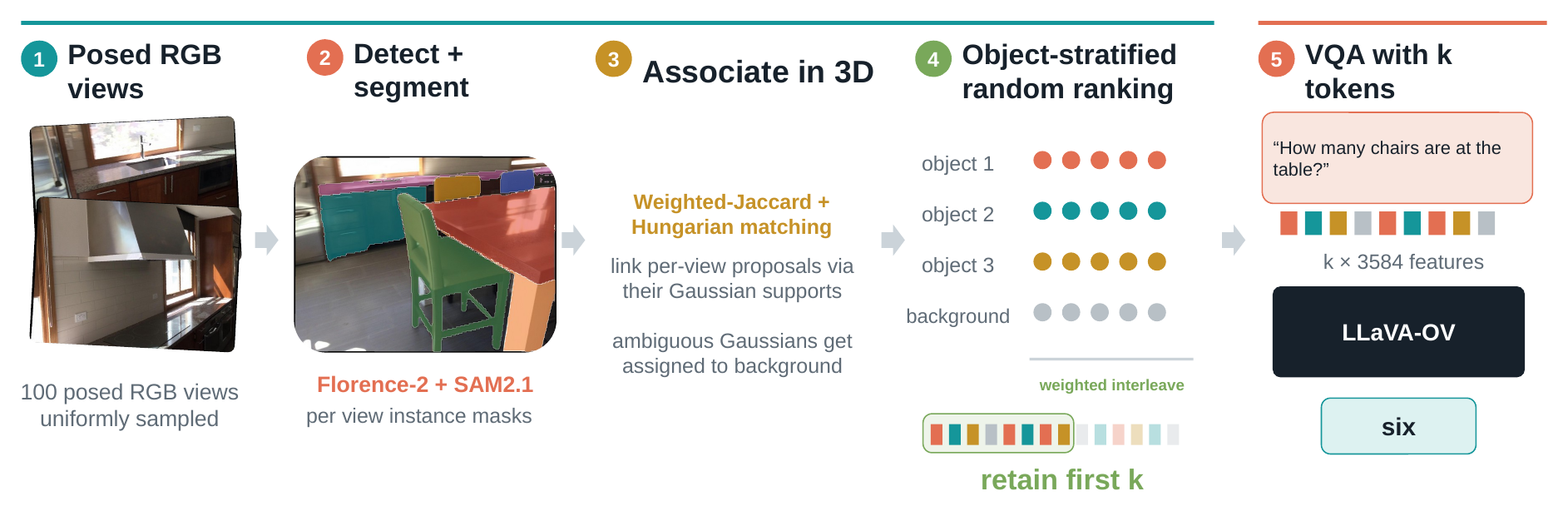}
    \caption{Overview of our object-based Gaussian sparsification pipeline. From up to 100 uniformly spaced finite-pose RGB views, Florence-2 and SAM2.1 produce per-view object masks. Gaussian alpha-compositing contributions project these proposals  into 3D, where weighted-Jaccard and Hungarian association form detected object tracks; ambiguous and structural Gaussians enter a shared background pool. Per-pool random orders are combined using equal foreground-track allocation and a 30\% background weight, to produce one nested scene ranking. The first $k$ Gaussian features are decoded by the autoencoder into $k$ 3584-dimensional LLaVA-OneVision visual tokens used for VQA.}
    \label{fig:detection}
\end{figure*}


\section{Related Work}
\label{sec:related}

\paragraph{3D visual question answering models}
ScanQA~\cite{azuma2022scanqa} introduced free-form question answering over reconstructed indoor scenes and established a point-cloud-based benchmark for spatial scene understanding.
SQA3D~\cite{ma2023sqa3d} extended this setting by conditioning questions on an agent's situated pose and orientation.
Subsequent 3D large multimodal models connect geometric scene representations to language models.
3D-LLM~\cite{hong20233dllm} injects learned 3D features into a Large Language Model (LLM) for a range of grounded language tasks, while LEO~\cite{huang2024leo} trains an embodied generalist agent over object-centric 3D observations.
LLaVA-3D~\cite{zhu2024llava3d} instead augments a pretrained multimodal model with 3D-aware tokens derived from multi-view inputs.
These approaches demonstrate the value of explicit 3D structures, but typically introduce a dedicated 3D encoder, object proposal pipeline, or task-specific alignment stage.
SplatTalk~\cite{thai2025splattalk} takes a complementary route: it learns language features inside a feed-forward Gaussian-splatting representation and decodes Gaussian features directly into the visual-token space of LLaVA-OneVision~\cite{li2024llavaonevision}.
Our study starts from SplatTalk's published representation and asks how many of these Gaussian tokens are actually needed at inference time.

\paragraph{Language-embedded neural fields and Gaussian splatting}
Three-dimensional Gaussian Splatting (3DGS)~\cite{kerbl20233dgs} represents a scene with anisotropic Gaussian primitives and supports high-quality real-time rendering.
Feature 3DGS~\cite{zhou2024feature3dgs} attaches distilled semantic features to Gaussian primitives, allowing a 3D field to inherit representations from 2D foundation models.
LangSplat~\cite{qin2024langsplat} compresses CLIP features into a language-aware Gaussian field for open-vocabulary querying, and OpenGaussian~\cite{wu2024opengaussian} develops point-level open-vocabulary understanding over 3D Gaussians.
These methods primarily evaluate localization, segmentation, or open-vocabulary recognition.
SplatTalk~\cite{thai2025splattalk} differs by reconstructing free-form LMM visual features and using individual Gaussian features as inputs to a language model for 3D VQA.
Its published inference procedure ranks decoded Gaussian features by entropy and retains up to the LMM context limit.
We revisit this design under controlled token budgets and compare uncertainty-based ranking with feature-blind, geometric, semantic, and joint spatial-semantic alternatives.

\paragraph{Visual-token reduction}
Visual-token redundancy has motivated pruning and merging methods for large multimodal models.
FastV~\cite{chen2024fastv} removes low-attention visual tokens inside the language model after early transformer layers; VisionZip~\cite{yang2025visionzip} identifies dominant and contextual tokens before LLM inference and merges redundant visual content.
More recent question-aware methods~\cite{li2026crisp,oh2026anchorprune} select a compact relevance anchor and then recover complementary context, rather than treating relevance and diversity as a single ranking objective.
Token reduction has also begun to incorporate 3D structure.

Fast3D~\cite{huang2025fast3d} predicts global attention to prune object-centric tokens in 3D multimodal models, while Geo3DPruner~\cite{li2026geo3dpruner} removes redundant spatial-video tokens using cross-view geometry and voxel-level coverage.
SeGPruner~\cite{li2026segpruner} combines attention-based semantic saliency with a geometry-aware diversity stage for multi-view 3D question answering.
Lai et al.~\cite{lai2026seeingonce} instead prune multi-view tokens online by projecting observed patches into a shared voxel space and suppressing spatially repeated evidence.

Recent analysis of conventional multimodal language models has shown that
projected image tokens contain substantial semantic redundancy. Fan et
al.~\cite{fan2026visualtokens} partition visual tokens into sink, dead, and
alive categories and find that approximately 40\% of tokens carry little or
no image-specific semantic information. While this result concerns 2D patch
tokens, we investigate whether substantially stronger redundancy is present
in spatially grounded 3D Gaussian language fields.

GaussianVLM~\cite{halacheva2025gaussianvlm} trains a prompt-conditioned module that re-tokenizes SceneSplat features into 128 aggregated scene tokens, rather than retaining 128 original Gaussians. We instead study question-independent sparsification to isolate redundancy in the existing representation.

Our setting is distinct in two ways.
First, the candidates are unstructured Gaussian primitives whose features already lie in the LMM visual-token space, rather than image patches, video tokens, or object proposals. Second, our work deliberately keeps selection independent of the question, isolating the information retained by the 3D scene representation itself.

\section{Methodology}
\label{sec:method}

\subsection{Post-Hoc and Training-Time Sparsification}

Post-hoc sparsification measures the redundancy of the learned
representation and can be applied directly to existing checkpoints. 
Let a reconstructed scene contain a set of Gaussian primitives
$\mathcal{G}=\{g_i\}_{i=1}^{N}$, with a learned semantic feature
$\mathbf{z}_i \in \mathbb{R}^{d}$ associated with each primitive. Given a token
budget $k$, a selector produces an ordered subset
\begin{equation}
    \mathcal{S}_k = S(\mathcal{G}, k),
    \qquad
    \mathcal{S}_k \subseteq \{1,\ldots,N\},
    \qquad
    |\mathcal{S}_k| = k.
\end{equation}

We define the retention ratio as $\rho = k/N$. Unless stated otherwise, the
ranking is computed once per scene and independently of the question and its
answers. 

In the post-hoc setting, we begin with the complete, trained semantic Gaussian
field and keep all model and autoencoder weights frozen. We rank the Gaussian embeddings using object-based, random,
geometric, semantic, and joint spatial-semantic criteria and retain
the first $k$ entries. For a budget $k$, we decode the selected
features into $\mathbf{F}_k \in \mathbb{R}^{k \times 3584}$ and pass
exactly these $k$ embeddings to SplatTalk's direct-feature interface
in the selector-defined ranking order, without padding or token
duplication.

\begin{figure*}[t]
    \centering
\includegraphics[width=1\linewidth]{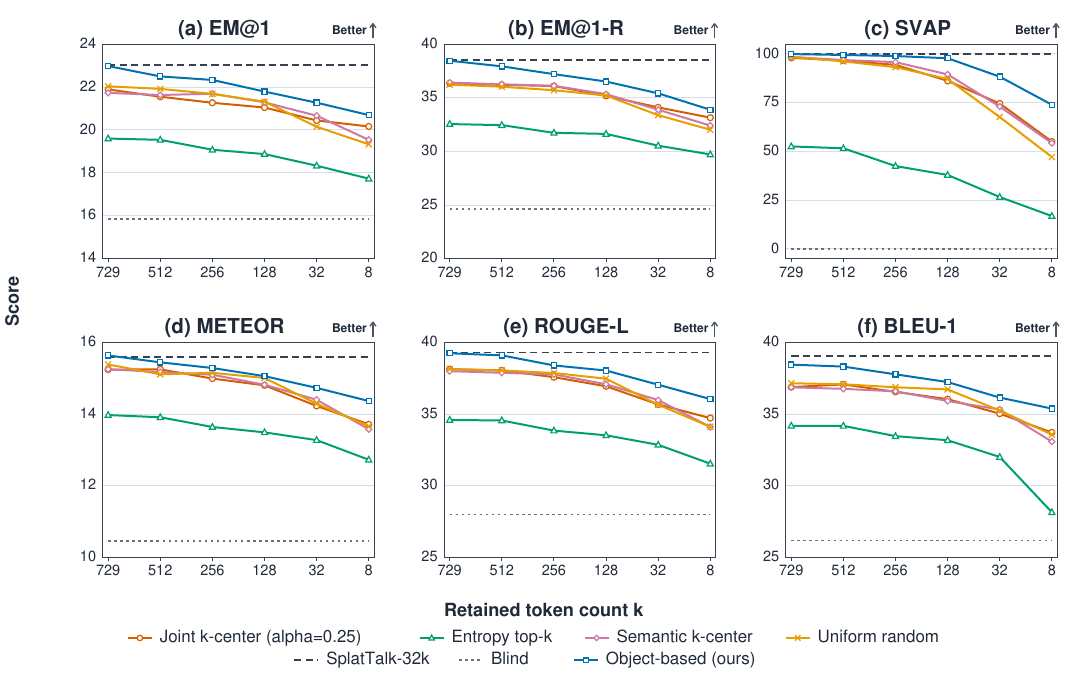}
    \caption{Performance on ScanQA when sparsifying from 729 down to 8 Gaussian embeddings, with the object-based selection offering the best results, maintaining performance down to 256 tokens. SplatTalk-32k denotes the original 32,076-token inference budget; Blind uses no visual tokens.}
    \label{fig:results_scanQA}
\end{figure*}

Prior work evaluates semantic Gaussian sampling at budgets of one or more
image-equivalent token blocks, with one block containing $27\times27=729$
tokens~\cite{thai2025splattalk}. However, our initial experiments show that
strong 3D VQA performance is retained even at substantially smaller budgets.
We therefore remove the 729-token block restriction and study the sub-block
regime from $k=8$ to $k=729$. This evaluation measures both the performance
curve and the point at which reducing the representation begins to remove
necessary scene information.

Finally, we transfer the best-performing selector to the training pipeline. In
this training-time setting, only the selected Gaussian primitives receive and
optimize semantic features. The remaining Gaussians continue to represent scene
appearance and geometry, but do not carry language embeddings. This provides additional benefits of reducing semantic-feature training cost by avoiding
the construction of a dense language field.

\subsection{Embedding selection}
\label{sec:selection}

Each selector constructs a deterministic ordered ranking
$\pi_s=(\pi_{s,1},\ldots,\pi_{s,729})$ for scene $s$.
A condition with budget $k$ consumes the exact prefix
$\{\pi_{s,1},\ldots,\pi_{s,k}\}$.

We evaluate the following ranking strategies.

\paragraph{Object-based} Global sampling methods may repeatedly select Gaussians belonging to the same
large object or structural surface while failing to represent smaller objects.
This is particularly limiting at low token budgets, where a small number of
redundant selections may remove an object from the retained representation
entirely. We therefore introduce object-stratified Gaussian sampling, which
distributes the available token budget across detected object tracks before
sampling within each track.

Following \cite{thai2025splattalk}, to reduce computational overhead, we uniformly select up to 100 finite-pose RGB observations from each scene as
nearby views are often redundant in image space. We first detect and
segment objects. We apply Florence-2-large using its generic
\texttt{<OD>} task prompt and use each detected bounding box to prompt
SAM~2.1 for an instance mask~\cite{xiao2024florence2,ravi2025sam2}.
No category list is supplied. We discard invalid,
low-confidence, and very small masks and suppress strongly overlapping
proposals. Detector labels are canonicalized by text post-processing. They are lowercased and normalized for punctuation and whitespace, then resolved to class identifiers from the published Open Images vocabulary \cite{kuznetsova2020openimages}.
Detections labeled as wall, floor, or ceiling are treated as structural
background rather than foreground objects. Overlapping mask pixels are
assigned deterministically using segmentation confidence, mask area,
canonical label, and proposal order.

For each detected proposal, we compute the contribution of every 3D Gaussian
to its mask using the Gaussian rasterizer. Let $M_{vr}(p)\in\{0,1\}$ indicate
whether pixel $p$ in view $v$ belongs to proposal $r$, and let $c_{vg}(p)$ denote the alpha-compositing contribution of Gaussian $g$
to pixel $p$ in view $v$. The proposal-to-Gaussian association mass is
\begin{equation}
    m_{vgr}
    =
    \sum_p M_{vr}(p)c_{vg}(p).
\end{equation}
Thus, a Gaussian receives high association mass when it contributes strongly
to pixels covered by a proposal mask.

Because the detector produces independent proposals in each image, proposals
must be associated across views. We represent each proposal by the smallest
set of Gaussian indices whose cumulative contribution accounts for 95\% of
its compositing mass, together with their normalized association weights.
Additional implementation details are provided in the supplementary material. We compare proposals
using the generalized weighted-Jaccard overlap of these sparse Gaussian
supports. Proposals are matched between views using deterministic one-to-one
Hungarian matching, with thresholds of $0.15$ for proposals having the same
canonical label and $0.40$ otherwise. Unmatched proposals initialize new
object tracks.

\begin{figure*}[t]
    \centering
\includegraphics[width=1\linewidth]{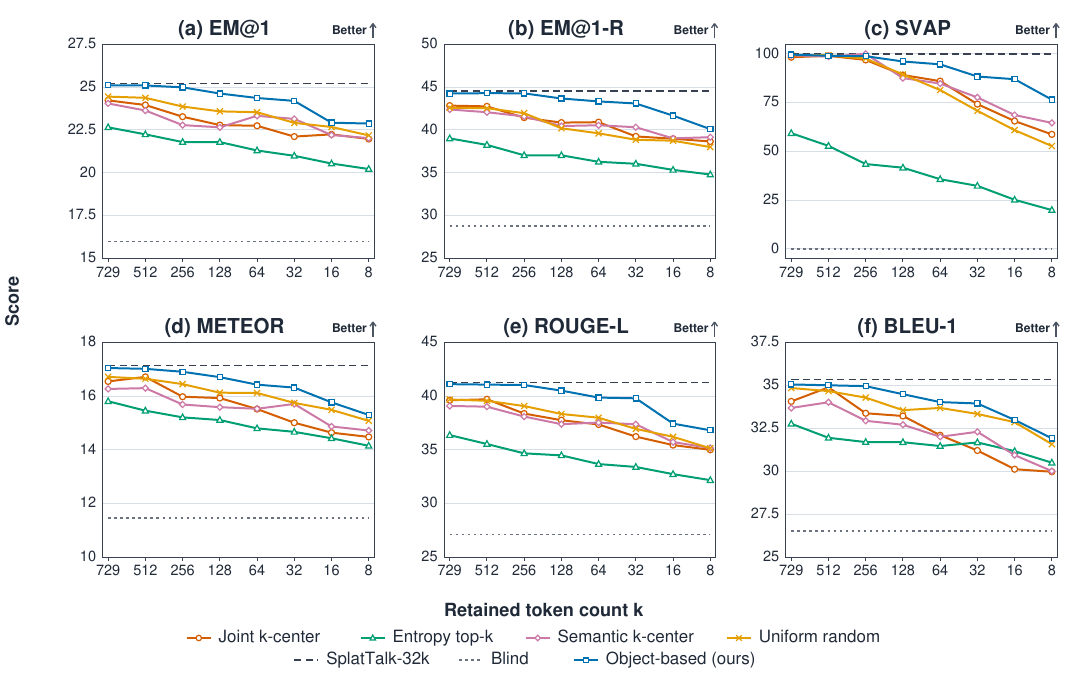}
    \caption{Sparsification from 729 to 8 embeddings on the MV-ScanQA dataset. We see modest performance drops in the 729 to 256 region across the metrics.}
    \label{fig:results_mv_scanQA}
\end{figure*}
After aggregating all proposals belonging to track $o$, we define its
association mass with Gaussian $g$ as
\begin{equation}
    C_{go}
    =
    \sum_{(v,r)\in o} m_{vgr}.
\end{equation}
The normalized association is
\begin{equation}
    a_{go}
    =
    \frac{C_{go}}
         {\sum_v\sum_p c_{vg}(p)+\epsilon}.
\end{equation}
Gaussian $g$ is assigned to the object track with the largest $a_{go}$ when
that score is at least $\tau_{\mathrm{assoc}}=0.5$. Ambiguous and unobserved
Gaussians are assigned to the background pool.

A direct allocation proportional to object size would cause large objects and
room surfaces to dominate the selected representation.

Instead, we combine uniform object coverage with sublinear size-based allocation. Let
$\mathcal{O}$ denote the set of foreground tracks and let $s_o$ denote the visible contribution of track $o$. Its normalized allocation weight is
\begin{equation}
    p_o
    =
    (1-\lambda)\frac{1}{|\mathcal{O}|}
    +
    \lambda
    \frac{s_o^\gamma}
         {\sum_{j\in\mathcal{O}}s_j^\gamma}.
\end{equation}

We assign a scheduling weight $q_{\mathrm{bg}}$ to the background pool. The
pool weights are therefore
\begin{equation}
    w_o=(1-q_{\mathrm{bg}})p_o,
    \qquad
    w_{\mathrm{bg}}=q_{\mathrm{bg}}.
\end{equation}

We use $\lambda=0.25$, $\gamma=0.25$, and
$q_{\mathrm{bg}}=0.3$. Thus, 30\% of the scheduling weight is assigned to
background, while the remaining 70\% is distributed among foreground tracks
using a mixture of 75\% uniform track coverage and 25\% sublinear
size-dependent allocation.
Within each object, we consider both random selection and farthest-point
selection; we find random selection to be the better of the two and use it
in most of the experiments.

\begin{table*}[t]
\centering
\caption{Resource scaling with the number of retained Gaussian tokens $k$ on the
ScanQA validation scenes. Inference-token retention is measured relative
to the $32{,}076$-token inference baseline, while dense-field retention is
computed as $71k/5{,}481{,}731$. The dense-field row uses the mean scene
size $\bar{N}_s=77{,}207$. Storage values are decimal MB and describe the
semantic Gaussian payload.}
\label{tab:resources}
\begin{tabular}{lccccc}
\toprule
\textbf{$k$} &
\textbf{\shortstack{Inference-token\\retention}} &
\textbf{\shortstack{Dense-field\\retention}} &
\textbf{\shortstack{Throughput\\(q/s)}} &
\textbf{\shortstack{Persistent selected\\payload (MB)}} &
\textbf{\shortstack{Decoded feature\\tensor (MB)}} \\
\midrule
$\bar{N}_s=77{,}207$ & --       & 100.000\% & --   & 83.08 & 553.42 \\
$32{,}076$           & 100.00\% & 41.545\%  & 0.58 & 34.50 & 229.92 \\
729                   & 2.27\%   & 0.944\%   & 10.2 & 0.79  & 5.23   \\
512                   & 1.60\%   & 0.663\%   & 12.3 & 0.56  & 3.67   \\
256                   & 0.80\%   & 0.332\%   & 14.3 & 0.27  & 1.84   \\
128                   & 0.40\%   & 0.166\%   & 16.7 & 0.14  & 0.92   \\
\bottomrule
\end{tabular}
\end{table*}

\paragraph{Uniform random}
Uniform random sampling is the simplest sparsification baseline, yet we see that it provides surprisingly robust performance. For each scene and global random seed, we derive a deterministic 64-bit seed by hashing the scene identifier and seed configuration with SHA-256. This seed initializes a pseudorandom permutation of all Gaussian indices, from which we retain the first \(k\) Gaussians for a token budget \(k\). Sampling is performed without replacement and is independent of Gaussian position, appearance, opacity, semantic features, and the question being answered.

\paragraph{Opacity top-$k$}
As a simple 3DGS-parameter baseline, we rank Gaussians by decreasing opacity $a_i$.
Opacity is available in the encoded Gaussian payload, so this method is independent from language embeddings. It tests whether primitives
with the strongest learned visual contribution are also the most useful for
VQA.

\paragraph{Farthest-point sampling}
FPS operates only on min-max-normalized Gaussian centers. The first Gaussian
is the one farthest from the scene centroid. Each subsequent Gaussian
maximizes its minimum Euclidean distance to the selected set. FPS therefore
promotes geometric coverage but does not use the learned language features.

\paragraph{Semantic $k$-center}
We $\ell_2$-normalize the 256-dimensional Gaussian features and choose one representative per occupied voxel: the member closest in cosine distance to that voxel's semantic centroid.
Starting from the representative nearest the global semantic centroid, greedy $k$-center repeatedly selects the candidate with the largest minimum cosine distance to the current set.

\paragraph{Decoded-feature entropy}
This mimics the selection score in \cite{thai2025splattalk}. For a decoded feature
$\mathbf{y}_i\in\mathbb{R}^{3584}$, we compute
\begin{equation}
    \mathbf{q}_i=\operatorname{softmax}(\mathbf{y}_i),\qquad
    H_i=-\sum_{c=1}^{3584}q_{ic}\log(q_{ic}+10^{-8}),
    \label{eq:entropy}
\end{equation}
and rank rows by decreasing $H_i$. Unlike the two $k$-center methods,
entropy scores each Gaussian independently and does not enforce spatial or
semantic coverage.

\paragraph{Joint spatial-semantic $k$-center}
This selector uses the same voxel representatives and initialization as
semantic $k$-center. For normalized centers
$\widetilde{\mathbf{p}}_i\in[0,1]^3$, we define
\begin{equation}
    d_{\mathrm{sp}}(i,j)
    = \frac{\lVert\widetilde{\mathbf{p}}_i-
    \widetilde{\mathbf{p}}_j\rVert_2}{\sqrt{3}}.
\end{equation}
At step $t$, each unselected candidate receives the score
\begin{equation}
\begin{split}
    r_t(i) ={}& \alpha
      \min_{j\in\mathcal{S}_{s,t-1}}d_{\mathrm{sp}}(i,j) \\
      &+(1-\alpha)
      \min_{j\in\mathcal{S}_{s,t-1}}d_{\mathrm{sem}}(i,j),
\end{split}
\end{equation}

and the candidate with maximum $r_t(i)$ is selected. We report the tested
setting $\alpha=0.25$, which places greater weight on semantic diversity while
retaining an explicit spatial-coverage term.

\subsection{Evaluation}
\label{sec:protocol}

We report EM@1~\cite{azuma2022scanqa},
EM@1-Refined~\cite{thai2025splattalk},
METEOR~\cite{banerjee-lavie-2005-meteor},
ROUGE-L~\cite{lin-2004-rouge}, and
BLEU-1~\cite{papineni-etal-2002-bleu}, together with inference
throughput and storage requirements.

Since we find that some of the questions in benchmarks can be answered in the blind or text-only mode, we present performance on questions not answered correctly by the blind model using Scaled Visually Attributable Performance (SVAP), defined as
\begin{equation}
    \operatorname{SVAP}(k)
    =
    100
    \frac{\sum_i (1-b_i)s_{k,i}}
         {\sum_i (1-b_i)f_i}.
\end{equation}
where $b_i$, $f_i$, and $s_{k,i}$ indicate whether question $i$
is answered correctly by the text-only, full-context, and $k$-token models, respectively.

\section{Results}

\subsection{Experimental setup}

We conduct our primary experiments on ScanQA
\cite{azuma2022scanqa} and MV-ScanQA \cite{mo2025mvscanqa}, using
their respective ScanNet validation splits. The ScanQA evaluation set
contains 4,675 questions across 71 scenes, while the MV-ScanQA
validation set contains 2,230 questions across 66 scenes. MV-ScanQA
places greater emphasis on questions that require evidence to be
integrated across multiple viewpoints, complementing the predominantly
single-view-solvable questions in ScanQA.

We use a model fine-tuned on the ScanQA training set for both datasets to further test generalizability. Because the model is not fine-tuned on MV-ScanQA, its performance on
this benchmark measures how well the learned representation transfers to
more explicitly multi-view reasoning tasks.

We also evaluate on \cite{huang2025beacon3d}, an object-centric benchmark designed to identify visual ignorance and inconsistencies between grounding and question
answering. 

\begin{table}[t]
\centering
\caption{Dataset statistics for ScanQA and MV-ScanQA.
\textit{Scenes} denotes the number of Gaussians with semantic embeddings
per scene in the vanilla setting; \textit{Objects} denotes the number of
ground-truth foreground objects per scene. Gaussian counts are reported
in thousands (K).}
\label{tab:dataset_stats}

\small
\setlength{\tabcolsep}{4pt}
\begin{tabular*}{\linewidth}{@{\extracolsep{\fill}}lrrrrr@{}}
\toprule
\textbf{Quantity} & \textbf{N} & \textbf{Mean} &
\textbf{Std. Dev.} & \textbf{Min} & \textbf{Max} \\
\midrule
ScanQA Scenes
    & 71 & 77.2K & 6.87K & 60.1K & 89.5K \\
MV-ScanQA Scenes
    & 66 & 77.2K & 7.08K & 60.1K & 89.5K \\
ScanQA Objects
    & 71 & 27.17 & 17.49 & 3 & 101 \\
MV-ScanQA Objects
    & 66 & 28.55 & 17.37 & 3 & 101 \\
\bottomrule
\end{tabular*}
\end{table}

Some of the questions in the benchmarks can potentially be answered without any visual information. We find that the blind-baseline EM@1-R scores are 24.61 and 28.77 for ScanQA and MV-ScanQA, respectively, compared to 38.52 and 44.52 achieved under full embeddings.

\subsection{Post-Hoc Sparsification}

\cref{fig:results_scanQA} and \cref{fig:results_mv_scanQA} compare post-hoc sparsification strategies on ScanQA and MV-ScanQA. Across both datasets, object-based selection obtains strong results at every evaluated token budget and across all six answer-quality metrics. This supports the hypothesis that explicitly distributing the token budget across object instances preserves more useful scene information than global sampling or feature-based ranking. 

\begin{figure}[h]
    \centering
\includegraphics[width=0.9\linewidth]{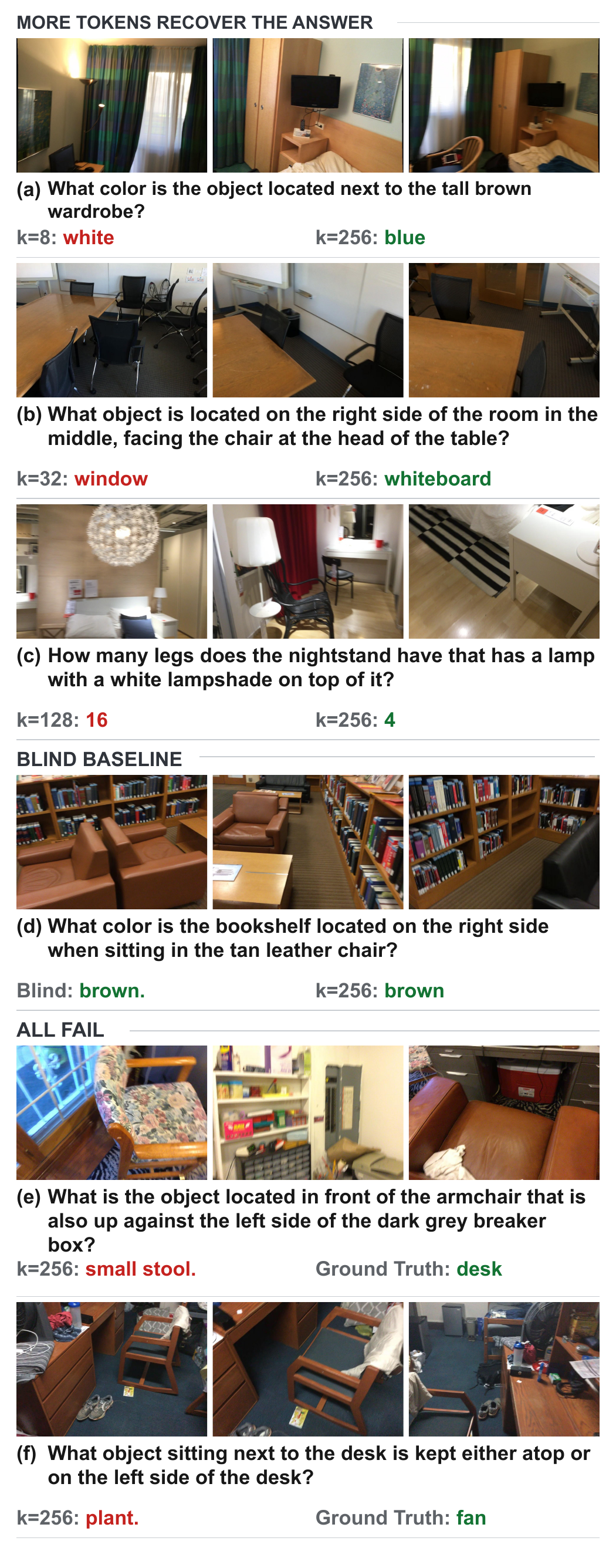}
    \caption{Samples where inclusion of more tokens leads to improvements in the answer quality (a-c), where even blind inference guesses the correct answer (d), and where all sparsification levels fail (e-f). Ground truth or answers evaluated as correct are denoted in green, and incorrect in red.}
    \label{fig:example_questions}
\end{figure}

METEOR, ROUGE-L, BLEU-1, and EM@1 scores follow the same trend, with minimal performance drop when decreasing from 729 to 256 embeddings. The trend is even more pronounced on MV-ScanQA, where EM@1-R stays within 0.1 points from 729 to 256 tokens, with a slight drop from 44.24 to 43.08 at 64 embeddings, and a falloff after 32. Uniform random sparsification also shows surprisingly strong performance, achieving higher scores than entropy top-k and joint k-center on MV-ScanQA in the 16 to 256 region on METEOR, ROUGE-L and BLEU-1. Entropy top-k shows the largest drop across the datasets. 

We found early in our experimental testing that 3DGS-parameter-based opacity top-k and FPS significantly underperform random selection, achieving only 30.29 - 33.42 and 31.12 - 33.42 EM@1-R respectively in the 8 to 729 region on a 1,188-question random ScanQA subset, so we dropped them from subsequent experiments.

Entropy-based ranking consistently underperforms the other selectors, despite requiring access to all densely decoded features. Together, these findings indicate that decoded-feature uncertainty is not a reliable proxy for question-independent scene coverage, while explicitly preventing small objects from disappearing from the retained representation provides a more robust selection criterion.

The modest score changes between 256 and 729 tokens accompany substantial downstream savings, as shown in \cref{tab:resources}. At 256 only 0.80\% of the 32,076-token baseline representation is retained with stable EM@1 and EM@1-R scores. This provides a 24.7x speedup compared to full context length and 40\% speedup over 729 tokens, while persistent encoded-feature storage decreases from 34.50 MB to 0.27 MB and decoded-feature memory decreases from 229.92 MB to 1.84 MB. 

Reducing the budget to 128 tokens further increases throughput to 16.7 questions per second, but the quality curves begin to decline more noticeably.
\subsection{Question answering capacity}
We examine how correlated the successfully answered questions are between the sparsification methods. We also examine the extent to which questions answered correctly at lower \(k\) remain correctly answered at higher \(k\). We calculate the mean pairwise Jaccard similarity among the sparsification approaches, with the mean at each level reported in \cref{tab:jaccard_k}.

\begin{table}[ht]
\centering
\caption{Mean pairwise Jaccard similarity across different values of $k$ across sparsification approaches.}
\label{tab:jaccard_k}
\begin{tabular}{c|cccccc}
\hline
$k$ & 8 & 32 & 128 & 256 & 512 & 729 \\
\hline
\% & 78.9 & 76.4 & 76.4 & 76.8 & 77.0 & 77.2 \\
\hline
\end{tabular}
\end{table}

When comparing different levels of sparsification, we find that \(
\frac{
\left|\text{correct at lower } k \;\cap\; \text{correct at higher } k\right|
}{
\left|\text{correct at lower } k\right|
}
\) for nearby k values tested is greater than 96\% across methods, but when comparing 32 to 729 or 16 to 729 the average drops to 82.4\%. The object-based sparsification performs better at 83.75\%, some examples can be seen in \cref{fig:example_questions}.

\begin{table}[h]
\centering
\caption{training-time performance on BEACON3D using the dense semantic field and [5\% retained / 95\% removed]}.
\label{tab:finetune}
\begin{tabular}{lcc}
\hline
\textbf{k} & \textbf{EM@1} & \textbf{EM@1-R} \\
\hline
Full & 25.1 & 41.2 \\
5\%  & 25.2 & 40.5 \\
\hline
\end{tabular}
\end{table}

\subsection{Fine-tuning}

We use object-based sparsification during the fine-tuning process by enforcing the selected k value. We only test Beacon3D at 5\% sparsification \cref{tab:finetune} as a pilot experiment due to computational limitations.

\subsection{Ablation}
\label{subsec:ablation}

\begin{table}[h]
\centering
\caption{EM@1 difference between random and within-pool FPS ordering in the object-based selector on ScanQA. Positive values favor random ordering.}
\label{tab:ablation}
\begin{tabular}{cc}
\hline
\textbf{k} & \textbf{Random $-$ FPS} \\
\hline
8   & -0.53 \\
32  &  0.14 \\
128 &  0.38 \\
256 &  0.38 \\
512 &  0.39 \\
729 &  0.42 \\
\hline
\end{tabular}
\end{table}

We compare random and farthest-point sampling within each object pool. FPS is better at the most extreme eight-token budget by 0.53 EM@1, but random sampling is better by 0.14–0.42 points from 32 through 729 tokens. The differences are consistently below one EM@1 point. We therefore use random within-object sampling in the main experiments because it is simpler and performs slightly better across the practically relevant 32 to 729 budget range. Further ablation details are provided in the supplementary material.

\section{Conclusion}

We present SparseTalk, a systematic study of sparsifying semantic embeddings in 3D Gaussian language fields for 3D visual question answering. Our experiments show that dense Gaussian language representations contain substantial redundancy: strong VQA performance can be retained with only a few hundred semantic embeddings per scene. Among the evaluated selection strategies, object-based sparsification achieves the highest observed aggregate scores under constrained token budgets, suggesting that maintaining coverage across object instances is more important than globally ranking Gaussians. In particular, performance remains largely stable around 256 retained tokens, despite using only 0.80\% of the 32,076-token inference baseline, while substantially reducing semantic-feature storage and decoded-feature memory and increasing inference throughput. We also find that uniform random sampling offers a surprisingly strong sparsification, outperforming other methods at most low embedding counts. These results indicate that explicit 3D semantic representations need not be dense to provide useful spatial grounding for downstream language reasoning.

Our study also exposes several limitations and directions for future work. First, the primary sparsification experiments are conducted on ScanQA and MV-ScanQA. A broader evaluation across datasets, scene types, and 3D language-field architectures is needed to establish the generality of the observed redundancy. Second, our object-based selector relies on 2D detection and segmentation followed by cross-view association to construct 3D object pools. Its effectiveness may consequently depend on the quality of these intermediate detections, particularly for small, occluded, or poorly segmented objects, and it introduces additional scene-level preprocessing. Finally, the presence of substantial blind-baseline performance makes it important for future evaluations to further study representation efficiency and dataset bias.

{
    \small
    \bibliographystyle{ieeenat_fullname}
    \bibliography{main}
}

\clearpage
\setcounter{page}{1}
\maketitlesupplementary

\begin{table*}[t]
    \centering
    \caption{ScanQA EM@1-Refined by sparsification method and token
    budget \(k\). Entries are mean \(\pm\) standard deviation; the best
    mean in each column is bold.}
    \label{tab:scanqa_em_at_1_refined}
    \resizebox{\textwidth}{!}{%
      \begin{tabular}{@{}lcccccc@{}}
        \toprule
        Sparsification method
          & $k=729$
          & $k=512$
          & $k=256$
          & $k=128$
          & $k=32$
          & $k=8$ \\
        \midrule
        Joint $k$-center ($\alpha=0.25$)
          & 36.33 $\pm$ 0.00
          & 36.19 $\pm$ 0.00
          & 36.09 $\pm$ 0.00
          & 35.18 $\pm$ 0.00
          & 34.10 $\pm$ 0.00
          & 33.14 $\pm$ 0.00 \\
        Object Based
          & \textbf{38.45} $\pm$ 0.08
          & \textbf{37.93} $\pm$ 0.55
          & \textbf{37.20} $\pm$ 0.53
          & \textbf{36.51} $\pm$ 0.13
          & \textbf{35.41} $\pm$ 0.18
          & \textbf{33.88} $\pm$ 0.35 \\
        Entropy top-$k$
          & 32.54 $\pm$ 0.00
          & 32.44 $\pm$ 0.00
          & 31.73 $\pm$ 0.00
          & 31.63 $\pm$ 0.00
          & 30.53 $\pm$ 0.00
          & 29.71 $\pm$ 0.00 \\
        Semantic $k$-center
          & 36.41 $\pm$ 0.00
          & 36.25 $\pm$ 0.10
          & 36.11 $\pm$ 0.00
          & 35.34 $\pm$ 0.00
          & 33.88 $\pm$ 0.00
          & 32.40 $\pm$ 0.00 \\
        Uniform random
          & 36.22 $\pm$ 0.85
          & 36.03 $\pm$ 0.98
          & 35.70 $\pm$ 0.34
          & 35.20 $\pm$ 0.18
          & 33.37 $\pm$ 0.48
          & 32.01 $\pm$ 0.51 \\
        \bottomrule
      \end{tabular}%
    }
  \end{table*}
  
\section{Repeatability of Sparsification and Decoding}
\label{app:reliability}

We examine how the results are affected by the randomness in the sparsification and decoding processes.

For stochastic sparsification methods such as uniform random sampling, variation arises from the scene-level selection seed and therefore from which Gaussians are retained. Methods such as joint and semantic k-center produce deterministic rankings; however, their results may still vary across repeated executions of the end-to-end pipeline. As a narrower diagnostic, we hold the exact decoded visual embeddings and their ordering fixed and vary only the model random seed, measuring any resulting variation in answers and metrics under greedy decoding.

\subsection{Model-Seed Sensitivity}
\label{sec:supp_model_seed_sensitivity}

Changing the seed may also affect the language model itself, making it unclear whether observed score variation originates from Gaussian selection or from answer generation. Although our inference protocol uses fixed weights and greedy decoding, nondeterministic GPU operations or seed-dependent model execution could still, in principle, alter predictions when token probabilities are close. We therefore isolate model-seed sensitivity while holding the complete visual input fixed.

We select 100 ScanQA validation questions from \texttt{scene0231\_00} and use the same ordered prefix of 729 visual tokens produced by object-based sparsification. Only the model seed is changed over $\{0,1,2,3,4\}$.

Across 500 generations and all ten pairwise seed comparisons, the prediction strings agree exactly for every question. Thus, under our fixed-weight greedy-decoding protocol, nominal model seeds introduce no measurable answer or metric variance when the visual tokens are unchanged.

\begin{figure*}[h]
    \centering
\includegraphics[width=1\linewidth]{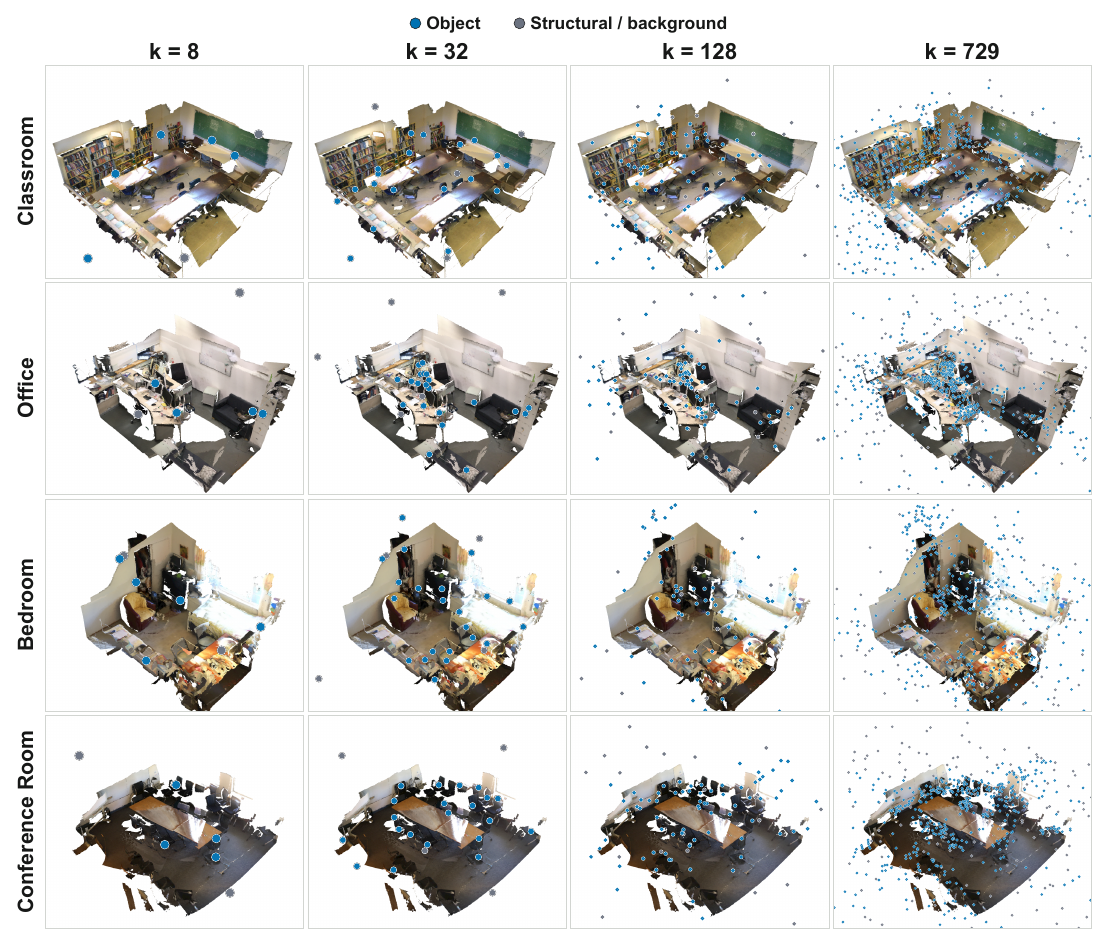}
    \caption{Visualization of the embedding locations at four k sparsification budgetsoverlayed on the 3d scene. Blue dots indicate the object allocations and the gray points indicate the backgorund/structural reserve. }
    \label{fig:grid}
\end{figure*}

\subsection{End-to-End Variance}

We measure repeatability on ScanQA questions. For each selector and token budget,
we execute the complete sparsification, decoding, and answer-generation pipeline
three times and report the mean and sample standard deviation of
$\mathrm{EM@1\text{-}R}$ \cref{tab:scanqa_em_at_1_refined} and SVAP \cref{tab:scanqa_svap}. Stochastic selectors use independent
scene-level selection seeds, whereas deterministic selectors reuse the same
ranking. Their variance therefore measures execution-level repeatability. Together with the fixed-input model-seed experiment above, these
measurements distinguish sensitivity to Gaussian selection from variability in
answer generation.

We see that standard deviation is highest for uniform random sampling, which is expected as it is the most stochastic method. The variance is notable for object-based sparsification when using random assignment within the objects, at 0.08 for $k=729$ and 0.35 for $k=8$, however the trends between the methods remain stable even when accounting for this.

\begin{table*}[h]
    \centering
    \caption{ScanQA SVAP by sparsification method and embedding budget \(k\).
    Entries are mean \(\pm\) standard deviation; the best mean in each
    column is bold.}
    \label{tab:scanqa_svap}
    \resizebox{\textwidth}{!}{%
      \begin{tabular}{@{}lcccccc@{}}
        \toprule
        Sparsification method
          & $k=729$
          & $k=512$
          & $k=256$
          & $k=128$
          & $k=32$
          & $k=8$ \\
        \midrule
        Joint $k$-center ($\alpha=0.25$)
          & 97.96 $\pm$ 0.00
          & 96.54 $\pm$ 0.00
          & 94.34 $\pm$ 0.00
          & 86.16 $\pm$ 0.00
          & 74.53 $\pm$ 0.00
          & 55.03 $\pm$ 0.00 \\
        Object-Based
          & \textbf{99.87} $\pm$ 1.14
          & \textbf{99.43} $\pm$ 2.88
          & \textbf{98.88} $\pm$ 2.01
          & \textbf{97.75} $\pm$ 2.62
          & \textbf{88.29} $\pm$ 1.75
          & \textbf{73.90} $\pm$ 4.62 \\
        Entropy top-$k$
          & 52.52 $\pm$ 0.00
          & 51.57 $\pm$ 0.00
          & 42.45 $\pm$ 0.00
          & 37.89 $\pm$ 0.00
          & 26.57 $\pm$ 0.00
          & 16.82 $\pm$ 0.00 \\
        Semantic $k$-center
          & 98.42 $\pm$ 0.00
          & 96.84 $\pm$ 0.33
          & 95.75 $\pm$ 0.00
          & 89.47 $\pm$ 0.00
          & 72.96 $\pm$ 0.00
          & 54.25 $\pm$ 0.00 \\
        Uniform random
          & 98.51 $\pm$ 3.02
          & 96.13 $\pm$ 5.08
          & 93.18 $\pm$ 2.06
          & 87.11 $\pm$ 0.66
          & 67.58 $\pm$ 2.49
          & 47.08 $\pm$ 5.97 \\
        \bottomrule
      \end{tabular}%
    }
  \end{table*}

\section{Models Used}
\label{sec:model_parameters}

We conduct most experiments on independently reproduced trainable components of SplatTalk~\cite{thai2025splattalk}: the visual-token autoencoder, the language-augmented Gaussian reconstruction model, and the ScanQA+SQA3D LoRA adapter. The pretrained SigLIP vision encoder~\cite{zhai2023siglip} and LLaVA-OneVision Qwen2-7B foundation model~\cite{li2024llavaonevision} were initialized from their public checkpoints and kept fixed except for the newly initialized LoRA parameters. After the release of the public models we compared the performance and found that the reproduced models achieved similar performance to the original models.

For each posed RGB observation, we extract LLaVA-OneVision visual tokens after the SigLIP encoder and multimodal projector. Each resulting $27\times27\times3584$ feature map is flattened into 729 token embeddings. We train a scene-general autoencoder with encoder dimensions
$3584\!\rightarrow\!2048\!\rightarrow\!1024\!\rightarrow\!512\!\rightarrow\!256$
and decoder dimensions
$256\!\rightarrow\!512\!\rightarrow\!1024\!\rightarrow\!2048\!\rightarrow\!2048\!\rightarrow\!3584$.
The encoder uses batch normalization and GeLU activations, and its 256-dimensional output is normalized to the unit hypersphere. Training minimized the sum of feature-space mean-squared error and cosine distance. We used a scene-disjoint training/validation split, a batch size of 256, AdamW with learning rate $10^{-4}$, and 100 epochs, retaining the checkpoint with the lowest validation reconstruction loss.

We train the SplatTalk reconstruction model on 500 ScanNet scenes~\cite{dai2017scannet} from the ScanQA training split. Following the FreeSplat architecture~\cite{wang2024freesplat}, the Gaussian decoder predicts the standard position, covariance, opacity, and appearance parameters together with a 256-dimensional language feature for every Gaussian. We use a parallel differentiable rasterizer. We use Adam with an initial learning rate of $10^{-4}$, a short linear warm-up, cosine decay, batch size one scene, and gradient clipping.

The trained autoencoder decoder maps each Gaussian's 256-dimensional feature back into the 3584-dimensional LLaVA token space. These decoded embeddings are packed as multimodal input tokens and paired with the combined ScanQA and SQA3D training annotations. We initialize a new rank-16 LoRA adapter~\cite{hu2022lora} with scaling factor 64 and dropout 0.05. LoRA updates are applied to the language model's query, key, value, and output projections and to its gate, up, and down MLP projections. The base language model and visual feature pipeline remain frozen.

We fine-tune for one epoch in bfloat16 precision on an NVIDIA H200 using AdamW, a peak learning rate of $10^{-5}$, 3\% linear warm-up, cosine decay, zero weight decay, gradient accumulation, and gradient checkpointing. Sequences use the Qwen conversation format and a maximum context length of 32,768 tokens. At evaluation time, we use the same Gaussian preprocessing, 44-block visual-token budget, prompt construction, and greedy decoding protocol for the reproduced and reference models.

\section{Object-Based Sparsification Details}
\label{sec:supp_object_sampling}

Here, we provide details of the detection, association, and ranking procedures for object-based sparsification. 

We numerically sort the posed RGB observations, discard frames with non-finite camera poses, and uniformly select up to 100 views across the remaining sequence. If fewer than 100 valid observations are available, we use all of them. We apply Florence-2-large with its generic \texttt{<OD>} task prompt, deterministic three-beam decoding, and no category list or question input. Each detected bounding box is then supplied to SAM~2.1 Hiera Large as an instance-segmentation prompt~\cite{xiao2024florence2,ravi2025sam2}.

We reject invalid boxes, boxes covering less than \(0.05\%\) or more than \(95\%\) of the image, masks with a SAM confidence below \(0.80\), and masks containing fewer than 64 pixels at the \(180\times320\) association resolution. Mask non-maximum suppression is applied at IoU \(0.80\) for proposals with the same canonical label and \(0.95\) otherwise. We retain at most 128 proposals per view.

The box-area limits exclude both extremely small detections and nearly full-image regions, while the SAM confidence and mask-area thresholds suppress uncertain or poorly supported masks. The same-label NMS threshold removes duplicate detections, whereas the higher cross-label threshold suppresses proposals with different labels only when their masks are nearly identical. 

Detector outputs are normalized using a fixed, scene-independent mapping to canonical object names derived from the Open Images vocabulary~\cite{kuznetsova2020openimages}. Normalization consists of lowercasing, punctuation and whitespace normalization, and fixed singular and synonym mappings. Labels without a mapped synonym retain their normalized detector name. Canonical labels are used only for cross-view association and structural-background identification. Detections corresponding to walls, floors, or ceilings are treated as structural background.

Before Gaussian association, the retained masks are converted into a disjoint proposal map. Pixels covered by multiple proposals are assigned deterministically according to segmentation confidence, mask area, canonical label, and original proposal order. Pixels not assigned to a retained foreground proposal belong to the contextual/background region.

Let \(\mathcal{P}_v\) denote the pixel domain of view \(v\), let \(M_{vr}(p)\in\{0,1\}\) indicate whether pixel \(p\) belongs to proposal \(r\), and let \(c_{vg}(p)\) denote the alpha-compositing contribution of Gaussian \(g\) to pixel \(p\). The contribution mass associating Gaussian \(g\) with proposal \(r\) in view \(v\) is
\begin{equation}
    m_{vgr}
    =
    \sum_{p\in\mathcal{P}_v} M_{vr}(p)c_{vg}(p).
    \label{eq:supp_proposal_mass}
\end{equation}

We obtain these masses directly from the differentiable Gaussian rasterizer. For each view, one auxiliary feature channel is assigned to each disjoint proposal region, and all per-Gaussian auxiliary features are initialized to zero. Let \(R_{vj}(p;F)\) denote rendered auxiliary channel \(j\) under per-Gaussian features \(F\). Since the scene geometry, opacity, and compositing weights are fixed, the rendered output is linear in \(F\). The gradient of the mask-weighted rendered channel with respect to Gaussian \(g\)'s corresponding auxiliary feature therefore equals \(m_{vgr}\). This gradient-based computation avoids materializing a dense Gaussian-by-pixel contribution tensor. We require all contribution masses to be finite and nonnegative. Because the disjoint proposal and background channels partition the rendered image, their summed contribution also recovers the corresponding all-pixel compositing mass.

Storing a dense Gaussian contribution vector for every proposal would be unnecessarily expensive. We therefore represent proposal \((v,r)\) by the smallest stable set of Gaussian indices \(\mathcal{S}_{vr}\) whose cumulative contribution accounts for at least \(95\%\) of its total mass, subject to a maximum of 2,048 Gaussians. If reaching \(95\%\) requires more than 2,048 Gaussians, we retain the 2,048 highest-contributing Gaussians. Each support's weights are normalized by the proposal's total contribution, quantized to \(10^{-6}\), and stored in Gaussian-index order.

Let \(u^A_g\) and \(u^B_g\) denote the sparse normalized weights of two proposals or tracks, with missing entries assigned zero weight. Their generalized weighted-Jaccard similarity is
\begin{equation}
    J(A,B)
    =
    \frac{\sum_g\min(u^A_g,u^B_g)}
         {\sum_g\max(u^A_g,u^B_g)}.
    \label{eq:supp_weighted_jaccard}
\end{equation}

The selected frames are processed in their original order. Proposals from the current frame are matched one-to-one with existing tracks by maximizing weighted-Jaccard similarity using Hungarian matching. A match is permitted when \(J\geq0.15\) for equal canonical labels or \(J\geq0.40\) when the labels differ. The stricter cross-label threshold accommodates variation in detector vocabulary while requiring stronger geometric evidence when canonical labels disagree. Unmatched proposals initialize new tracks.

After adding a proposal, we update the track's accumulated Gaussian contributions and recompute its sparse support. The track label is the canonical label receiving the largest accumulated visible contribution. Similarities are quantized before assignment, and stable proposal and track identifiers resolve exact ties. To bound computation, we retain at most 512 foreground tracks per scene, prioritized by accumulated visible contribution. Association mass outside this set is incorporated into the shared contextual/background pool.

In subsequent aggregation, omitted support entries are treated as zero, and retained normalized weights are rescaled by the proposal's total contribution mass. For notational simplicity, we continue to denote these reconstructed sparse masses by \(m_{vgr}\). After aggregating all proposals assigned to track \(o\), we define its association mass with Gaussian \(g\) as \(C_{go}=\sum_{(v,r)\in o}m_{vgr}\). We define the total visible contribution of Gaussian \(g\) across the selected views as \(V_g=\sum_v\sum_{p\in\mathcal{P}_v}c_{vg}(p)\). The normalized track association is
\begin{equation}
    a_{go}
    =
    \frac{C_{go}}{V_g+\epsilon}.
    \label{eq:supp_track_confidence}
\end{equation}

Gaussian \(g\) is assigned to \(\operatorname*{arg\,max}_o a_{go}\) when the winning association confidence is at least \(\tau_{\mathrm{assoc}}=0.5\). All remaining Gaussians, including structural and weakly associated regions, are collected in the shared contextual/background pool. Retaining this pool preserves scene layout and visual evidence not represented by a confident foreground track.

Let \(\mathcal{O}\) denote the set of confidently assigned foreground tracks, let \(\pi(g)\) denote the pool assignment of Gaussian \(g\), and define the visible size of track \(o\) as \(s_o=\sum_{g:\,\pi(g)=o}V_g\). We combine uniform track coverage with sublinear visible-size weighting:
\begin{equation}
    p_o
    =
    (1-\lambda)\frac{1}{|\mathcal{O}|}
    +
    \lambda
    \frac{s_o^\gamma}
         {\sum_{j\in\mathcal{O}}s_j^\gamma}.
    \label{eq:supp_object_weight}
\end{equation}

We reserve a fraction \(q_{\mathrm{bg}}\) of the scheduling weight for the contextual/background pool. Thus, each foreground track receives weight \(w_o=(1-q_{\mathrm{bg}})p_o\), while the background pool receives \(w_{\mathrm{bg}}=q_{\mathrm{bg}}\). For the configuration reported in the paper, we use \(\lambda=0.25\), \(\gamma=0.25\), and \(q_{\mathrm{bg}}=0.3\). Consequently, the foreground allocation combines \(75\%\) uniform track coverage with \(25\%\) fourth-root visible-size weighting, while the contextual/background pool receives \(30\%\) of the scheduling weight.

For the primary random variant, we construct a deterministic random permutation independently within every foreground and background pool. The random seed is derived from the scene identifier, experiment seed, and pool identifier, but not from the requested token budget. We combine these pool-specific orders using weighted deficit scheduling. At each step, every nonempty pool accumulates credit in proportion to its scheduling weight normalized over the currently active pools. We emit the next Gaussian from the pool with the largest credit and then subtract one from that pool's credit. When a pool is exhausted, its allocation is redistributed among the remaining pools. Stable scene-dependent hashes resolve exact scheduling ties.

For the farthest-point variant, the within-pool sequence begins with the Gaussian having the greatest visible contribution and then repeatedly selects the Gaussian with the largest minimum Euclidean distance from the previously selected points. The same weighted interleaving procedure is used for both within-pool selectors.

The random variant produces one immutable 729-Gaussian ranking per scene and seed, while the farthest-point variant produces one deterministic ranking per scene. A token budget \(k\) consumes exactly the first \(k\) entries, making all evaluated budgets nested prefixes of the same ranking. Only these \(k\) encoded Gaussian features are decoded into 3,584-dimensional LLaVA features. They are provided to the multimodal model with shape \(1\times3584\times1\times k\), which the existing LLaVA input preparation flattens into exactly \(k\) visual tokens.

\subsection{Allocation Parameter Ablation}
\label{sec:object_sampling_parameters}

We ablate the three parameters controlling object-based token allocation: the background scheduling weight $q_{\mathrm{bg}}$, the foreground size-mixture coefficient $\lambda$, and the size exponent $\gamma$. Foreground track $o$ receives normalized weight
\begin{equation}
p_o
=
(1-\lambda)\frac{1}{|\mathcal{O}|}
+
\lambda
\frac{s_o^{\gamma}}
{\sum_{j\in\mathcal{O}} s_j^{\gamma}},
\end{equation}
with final scheduling weights
\begin{equation}
w_o=(1-q_{\mathrm{bg}})p_o,
\qquad
w_{\mathrm{bg}}=q_{\mathrm{bg}}.
\end{equation}
Our default configuration is $q_{\mathrm{bg}}=0.3$, $\lambda=0.25$, and $\gamma=0.25$.

We evaluate parameter sensitivity on a fixed development subset of 1,188 ScanQA questions and report EM@1-R on embedding budget of 256.

\begin{table}[t]
\centering
\caption{Allocation parameter ablation on the fixed 1,188-question
ScanQA development subset at embedding budget \(k=256\). Each block
varies one parameter while holding the other two at their default
values. Bold indicates the default setting.}
\label{tab:allocation_parameter_ablation}
\begin{tabular}{lcc}
\toprule
Parameter & Value & Avg.\ EM@1-R \\
\midrule
$q_{\mathrm{bg}}$ & $0.0$  & $35.10$ \\
$q_{\mathrm{bg}}$ & $0.1$  & $37.10$ \\
$q_{\mathrm{bg}}$ & $0.3$  & $\mathbf{37.21}$ \\
$q_{\mathrm{bg}}$ & $0.5$  & $36.80$ \\
\midrule
$\lambda$ & $0.0$  & $37.02$ \\
$\lambda$ & $0.25$ & $\mathbf{37.21}$ \\
$\lambda$ & $0.5$  & $37.09$ \\
\midrule
$\gamma$ & $0.25$ & $\mathbf{37.21}$ \\
$\gamma$ & $0.5$  & $37.12$ \\
\bottomrule
\end{tabular}
\end{table}

The clearest effect comes from the background allocation. Removing it entirely, $q_{\mathrm{bg}}=0$, reduces average EM@1-R by approximately $2.11$ points relative to $q_{\mathrm{bg}}=0.3$, indicating that contextual and background Gaussians provide useful scene information beyond detected foreground objects. However, performance is relatively insensitive to the exact nonzero value: $q_{\mathrm{bg}}=0.1$ and $0.3$ perform similarly, while $q_{\mathrm{bg}}=0.5$ gives only a modest decrease.

The foreground allocation parameters have substantially smaller effects. A positive $\lambda$ provides a slight improvement over purely uniform object allocation, but performance between $\lambda\in\{0.25,0.5\}$ is similar. Likewise, changing $\gamma$ from $0.25$ to $0.5$ has negligible impact. Overall, the ablation suggests that preserving a nonzero background allocation is important, whereas the precise values of $\lambda$ and $\gamma$ are not critical within the tested ranges.

\end{document}